\documentclass[10pt,twocolumn,letterpaper]{article}

\usepackage{cvpr}              

\definecolor{our_results_color}{rgb}{0.95, 0.95, 0.95}
\definecolor{lime}{rgb}{0.2, 0.9, 0.2}

\definecolor{our_results_color}{rgb}{0.95, 0.95, 0.95}
\definecolor{lime}{rgb}{0.2, 0.9, 0.2}

\usepackage{adjustbox}
\usepackage{float}
\usepackage{afterpage}

\definecolor{cvprblue}{rgb}{0.21,0.49,0.74}
\usepackage[pagebackref,breaklinks,colorlinks,allcolors=cvprblue]{hyperref}
\usepackage[accsupp]{axessibility}  

\title{IAUNet: Instance-Aware U-Net}

\author{
Yaroslav Prytula$^{1,2}$ \quad Illia Tsiporenko$^{1}$ \quad Ali Zeynalli$^{1}$ \quad Dmytro Fishman$^{1,3}$\\
$^{1}$Institute of Computer Science, University of Tartu\\ $^{2}$Ukrainian Catholic University, \quad $^{3}$STACC OÜ, Tartu, Estonia\\
{\tt\small \{yaroslav.prytula, illia.tsiporenko, ali.zeynalli, dmytro.fishman\}@ut.ee}\\
\tt\small s.prytula@ucu.edu.ua
}

\begin{document}
\maketitle
\begin{abstract}
Instance segmentation is critical in biomedical imaging to accurately distinguish individual objects like cells, which often overlap and vary in size. Recent query-based methods, where object queries guide segmentation, have shown strong performance. While U-Net has been a go-to architecture in medical image segmentation, its potential in query-based approaches remains largely unexplored. In this work, we present IAUNet, a novel query-based U-Net architecture. The core design features a full U-Net architecture, enhanced by a novel lightweight convolutional Pixel decoder, making the model more efficient and reducing the number of parameters. Additionally, we propose a Transformer decoder that refines object-specific features across multiple scales. Finally, we introduce the 2025 Revvity Full Cell Segmentation Dataset \footnote{Dataset available at: https://github.com/SlavkoPrytula/IAUNet}, a unique resource with detailed annotations of overlapping cell cytoplasm in brightfield images, setting a new benchmark for biomedical instance segmentation. Experiments on multiple public datasets and our own show that IAUNet outperforms most state-of-the-art fully convolutional, transformer-based, and query-based models and cell segmentation-specific models, setting a strong baseline for cell instance segmentation tasks. Code is available at \url{https://github.com/SlavkoPrytula/IAUNet}

\end{abstract}
    
\section{Introduction}
\label{sec:intro}

Accurate cell instance segmentation is crucial in biomedical imaging \cite{artseg}, as it enables the precise identification and analysis of individual cells. This process is essential for understanding cellular behaviors and disease mechanisms~\cite{kherlopian2008review}. However, the diverse and irregular shapes of cells present significant challenges for segmentation algorithms~\cite{Moen2019-cg, Shrestha2023-pl}. Variations in cell morphology, overlapping structures, and differing imaging conditions can lead to segmentation errors~\cite{artseg}. Addressing these challenges requires the development of advanced segmentation models capable of handling the complexities associated with cell shapes. Deep learning models have driven substantial progress in cell segmentation, often surpassing traditional methods \cite{he2015delvingdeeprectifierssurpassing, ronneberger2015unet, unet++}. However, cell segmentation remains challenging due to heterogeneous cell appearances, overlaps, and varied object densities across different microscopy modalities, requiring models that generalize well across conditions.

Brightfield microscopy, valued for its simplicity and affordability, presents unique challenges for segmentation \cite{artseg}. Unlike fluorescence microscopy, which requires staining, and phase-contrast microscopy, which relies on specialized optics to enhance contrast in transparent specimens, brightfield uses natural light alone \cite{brightfiled_alternative_to_whole_cell_fluorescence}. This makes brightfield ideal for real-time observation in both research and clinical settings \cite{brightfield_multiplex, detecting_and_tracking_nonfluorescent, artseg}. However, brightfield images are inherently low-contrast, noisy, and variable, making precise cell segmentation difficult and underscoring the need for specialized approaches tailored to this modality.

Many previous works have adapted instance segmentation models from natural images to medical imaging without model-specific adjustments \cite{mask_rcnn, yolov8, yolov9}. In contrast to many of these methods, U-Net \cite{ronneberger2015unet} has long been a go-to architecture for semantic segmentation. Its lightweight framework, characterized by skip connections and an encoder-decoder structure, enables precise localization and the effective capture of intricate details, making it especially well-suited for biomedical applications. U-Net’s efficiency is particularly advantageous when working with smaller microscopy datasets, as it typically requires less data to train compared to more complex models. This is why we chose to focus on U-Net in our work, building on its established popularity and applicability to microscopy data. 

Building on the success of DETR \cite{detr} in object detection, query-based single-stage instance segmentation methods \cite{maskformer, mask2former, mask_dino, queryinst, fastinst, pctrans} have gained prominence. These methods move away from traditional convolutional approaches, utilizing the powerful attention mechanism \cite{attention_is_all_you_need} together with learnable queries to directly predict object classes and segmentation masks in an end-to-end fashion. However, these models typically rely on single-level features to generate queries, refining them without leveraging the full range of features available from skip connections and decoder feature maps. 
This limits their ability to capture the rich multi-scale context necessary for precise instance refinement.

To address these limitations, we bridge the gap between the U-Net model, widely used in biomedical imaging, and the task of instance segmentation. We present IAUNet, a novel architecture that enhances U-Net with instance-awareness through query-based mechanisms. This design incorporates a lightweight convolutional Pixel decoder, enabling the model to scale effectively with larger backbones while maintaining strong performance across both small and large datasets. IAUNet also introduces a Transformer decoder for multi-scale object feature refinement.

As part of our contributions, we introduce the 2025 Revvity Full Cell Segmentation Dataset, specifically designed for benchmarking model performance. The dataset includes hundreds of carefully annotated cell instances in high-resolution brightfield images, each thoroughly hand-labeled and validated. One of its unique features is the precise annotation of cell borders, even in cases of overlapping cells, allowing it to capture complex cell interactions. This dataset is a valuable resource for evaluating model accuracy in capturing fine details and handling challenging segmentation tasks with intricate cell morphologies. 

Our main contributions are as follows:

\begin{itemize}[leftmargin=2.3em]
    \item[--] We introduce a lightweight Pixel-Transformer decoder within U-Net for multi-scale object feature refinement, efficiently scaling with larger backbones. 
    \item[--] We introduce a novel 2025 Revvity Full Cell Segmentation Dataset with detailed annotations and provide a benchmark for instance segmentation. 
\end{itemize}

\section{Related Work}
\label{sec:related_work}

Instance segmentation methods are generally categorized into region-based, query-based, and specialized approaches that often require preprocessing. 

\noindent \textbf{Region-based Methods} exemplified by Mask R-CNN \cite{mask_rcnn, faster_rcnn, roipool}, have set a standard in natural image segmentation with their proposal-based structure.
Building on Faster R-CNN \cite{faster_rcnn}, Mask R-CNN adds a mask prediction branch for end-to-end instance segmentation by first detecting bounding boxes and then applying Region of Interest (RoI) operations like RoI-Pooling \cite{roipool} or RoI-Align \cite{mask_rcnn} to extract features for classification and mask generation. However, these two-stage methods often generate numerous redundant region proposals, reducing efficiency \cite{fastinst, sparseinst}. Although they perform well on many benchmarks, their reliance on small RoI regions frequently leads to coarse mask predictions. Some methods focus on enhancing the precision of detected bounding boxes \cite{cascade_rcnn}, while others, like PointRend \cite{pointrend}, specifically address low-quality segmentation masks by refining boundaries at uncertain points to improve segmentation quality. However, even with these advancements, traditional region-based methods face limitations in biomedical image segmentation \cite{pctrans}, where objects have complex shapes, orientations, and sizes. In these settings, traditional axis-aligned bounding boxes struggle to capture detailed contours, particularly for irregular and overlapping cellular structures \cite{oriented_boxes_for_accurate_instance_segmentation, instance_cut}.

\noindent \textbf{Specialized Cell Instance Segmentation Methods} like StarDist \cite{stardist} segment biomedical images by representing objects as star-convex polygons, predicting distances from a central point to boundaries in multiple directions. This method, along with other similar approaches like DeepWatershed \cite{deepwatershed} and Micro-Net \cite{micro_net}, works well for star-shaped or rounded cells but struggles with irregular, elongated shapes and overlapping cells. CellPose \cite{cellpose}, by contrast, similar to Hover-Net \cite{hovernet}, uses a U-Net to predict horizontal and vertical gradients alongside a binary cell map, creating a vector field that directs pixels toward the cell center. While this method effectively separates individual cells, it often relies on an additional size model \cite{cellpose2} to estimate object diameters, which becomes challenging with varying cell sizes and shapes. Although these methods offer advancements over traditional techniques, they remain limited in accurately segmenting overlapping cells and handling complex cellular morphologies.

\begin{figure*}[t!]
    \centering
    \includegraphics[width=\textwidth]{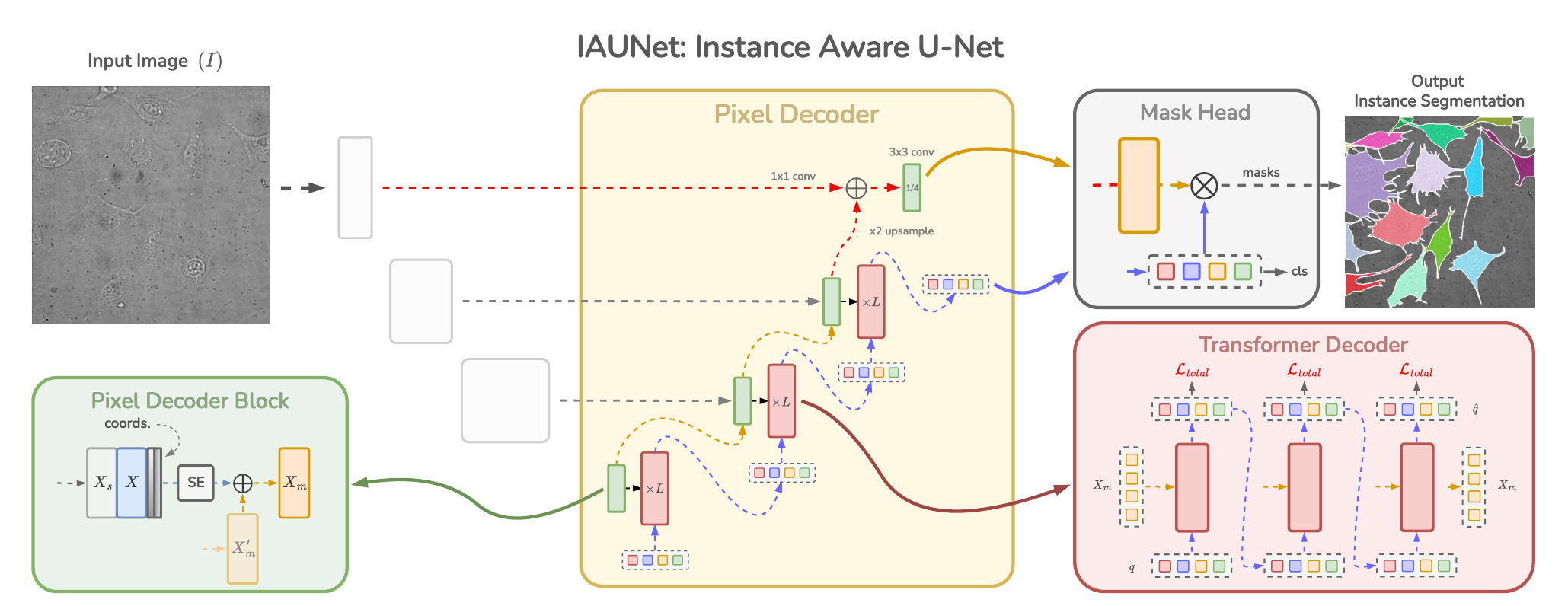}
    \caption{\textbf{Model overview.} Overview of the IAUNet architecture, highlighting the Pixel and Transformer Decoder stages. Given an input image \( I \), the encoder extracts multi-scale features as skip connections for the Pixel decoder. At each decoder block, we add skip connections \( X_s \) to the main features \( X \) and inject normalized coordinate features for CoordConv. Stacked depth-wise convolutions with an SE block refine spatial information, generating mask features \( X_m \). The Transformer decoder then processes learnable queries \( q \) through three Transformer blocks per layer, iteratively refining them with \( X_m \). Deep supervision loss is applied after each Transformer block using updated queries \( \hat{q} \) and high-resolution mask features.}
    \label{fig:iaunet_main}
\vspace{-0.4cm}
\end{figure*}

\noindent \textbf{Query-based Methods} have gained popularity since the introduction of DETR \cite{detr}, which demonstrated the potential of Transformer-based architectures for instance segmentation. Unlike traditional region-based models, query-based methods use object queries to directly predict object instances, removing the need for predefined bounding boxes. Building on DETR, models like Mask2Former~\cite{mask2former} and FastInst~\cite{fastinst} introduced masked attention to improve convergence and segmentation precision. These models heavily rely on producing fine features using MSDeformAttn Transformer \cite{msdeformattn} Pixel decoder. MaskDINO~\cite{mask_dino} further advances instance segmentation by adding a mask prediction branch that generates high-resolution binary masks through query embeddings for unified segmentation tasks. Recently, adaptations of query-based models have also emerged in the biomedical domain. For example, Cell-DETR~\cite{cell_detr} adapts DETR specifically for cell segmentation by leveraging queries to detect individual instances. The model uses the final feature map of the encoder for query initialization, limiting multi-scale query refinement across decoder features. Its segmentation head applies multi-head attention between encoder and decoder features, followed by a CNN decoder. However, it merges queries with decoder features only at the lowest layer, forcing the CNN decoder to handle most of the instance separation. This makes the model inefficient for high-resolution inputs with many queries. Additionally, Cell-DETR applies softmax to suppress overlapping predictions, reducing its ability to segment occluding cells effectively. Recent work, such as PCTrans~\cite{pctrans}, built on Mask2Former, introduces a position-guided transformer with a query contrastive loss. Similar to DETR, position guidance is done by predicting the normalized center coordinates of each object. While natural objects are often convex, cells present more complex shapes, with centers that often fall outside boundaries, particularly in elongated structures \cite{omnipose}, making mask representation less effective. 

All previous query-based models \cite{mask_dino, mask2former, detr, pctrans} have been designed around the idea of a Transformer-based Pixel decoder, which raises concerns about scalability to smaller datasets. Unlike these models, we propose a lightweight Pixel decoder that improves performance on smaller datasets. In \cref{table:metrics_all_revvity}, we show that IAUNet consistently outperforms state-of-the-art models across different backbones while maintaining strong results on large-scale datasets (\cref{table:metrics_all}). Our experiments show that IAUNet outperforms most alternatives while using fewer parameters and achieving higher efficiency.

\section{Model Overview}
\label{sec:model_overview}

The IAUNet model follows a U-Net design, illustrated in \cref{fig:iaunet_main}. The model consists of three main components: an encoder, a Pixel decoder, and a Transformer decoder. Given an input image $I \in \mathbb{R}^{H \times W \times 3}$, the encoder produces four multi-scale semantic feature maps at resolutions of $1/4$, $1/8$, $1/16$, and $1/32$ relative to the original image. These feature maps are utilized as skip connections in the decoder. The Pixel decoder first processes these features to generate the main decoder features $X$. At each decoder layer, these features pass through a lightweight mask branch to produce refined mask features, $X_m$, which then interact with object queries. The Transformer decoder further refines instance queries with mask features. This process is iterative, with updated queries passing through each decoder stage. In the final stage, the mask head combines mask features and instance queries to produce output instance masks.

\subsection{Pixel Decoder}
\label{sec:pixel_decoder}

In the biomedical domain, U-Net \cite{ronneberger2015unet}, with all its variants \cite{unet++, swin_unet, transunet, unetr}, still holds the ground as the most superior network for accurate segmentation. This is primarily due to the design of U-Net's decoder, which maintains high semantic consistency through the use of skip connections. We include a convolutional decoder, referred to as the Pixel decoder. Our Pixel decoder (\cref{fig:iaunet_main}, middle panel) works with two feature types: main features $X$ and mask features $X_m$. The main features serve a similar role to those in the vanilla U-Net, aggregating spatial context across the image using skip connections $X_s$. The mask features refine $X$ and capture richer semantic information. All these features are specifically designed to support instance segmentation and are tightly integrated with the Transformer decoder (see \cref{sec:transformer_decoder}).

\begin{figure*}[t!]
    \centering
    \includegraphics[width=\textwidth]{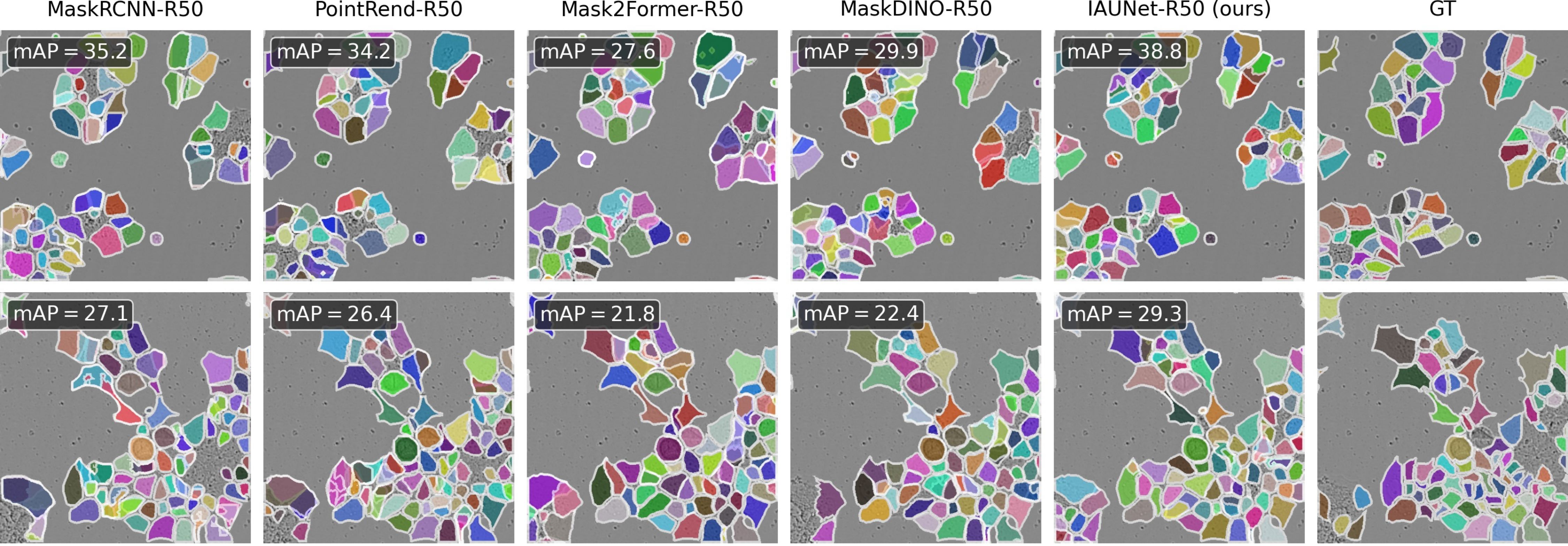}
    \caption{\textbf{LIVECell}. Visualization of instance segmentation predictions on the LIVECell dataset across different state-of-the-art models (using R50 backbone). We also report per-image AP score. Last columns shows ground-truth annotations.}
    \label{fig:main_predictions_livecell}
\vspace{-0.2cm}
\end{figure*}

\vspace{-0.6cm}
\begin{flalign}
&\! X = SE \Big( G_x \big( [X_s, X'] \big) + X' \Big) \label{eq:main_feats_update} \\ 
&\! X_m = G_m \Big( X_m' + X \Big) \label{eq:mask_feats_update}
\end{flalign}

At each level, the corresponding skip connection $X_s$ is first mapped to a 256-dimensional feature map. Then it gets concatenated with the upscaled decoder features $X'$ from the previous layer and passed through a lightweight double $3 \times 3$ point-wise convolution, batch normalization, and ReLU layer $G_x$ (\cref{eq:main_feats_update}). Next, we apply a Squeeze-and-Excitation (SE) \cite{squeeze_and_excitation} block to produce the final main features $X$. Next, we update mask features by adding the main features $X$ and the upscaled mask features $X_m'$ from the previous layer followed by two stacked $3 \times 3$ convolutional layers $G_m$ \cref{eq:mask_feats_update}. The whole process preserves multi-scale semantic information while maintaining a lightweight structure. The updated mask features are then used for query refinement in the corresponding Transformer blocks. Finally, we use bilinear upscaling to propagate all features to the next decoder layer.

\subsection{Transformer Decoder}
\label{sec:transformer_decoder}

Object queries are central to instance segmentation \cite{solq, queryinst, mask2former, fastinst}, serving as learnable embeddings that represent each object as a unique $D$-dimensional feature vector. These queries group pixel features relevant to each specific object, typically through a cross-attention mechanism. They are particularly important in Transformer architectures \cite{detr}, where they are processed and refined in an end-to-end manner. In existing models such as DETR \cite{detr}, Deformable DETR \cite{deformable_detr}, MaskFormer \cite{maskformer}, and Mask2Former \cite{mask2former}, queries are central to representing objects for segmentation or detection tasks. In our work, we use $N$ learnable queries $q \in \mathbb{R}^{N \times 256}$. Each query is thus a 256-dimensional representation, capturing the finer semantic object features. These instance queries are progressively refined with mask features $X_m$ through a multi-layer Transformer decoder (see \cref{sec:transformer_decoder}). At each decoder layer $l \in [1, L]$, we use three Transformer decoder layers. Queries from the previous decoder layer are iteratively processed through these layers (\cref{fig:iaunet_main}, red block) with the corresponding flattened mask features $X_m \in \mathbb{R}^{L \times 256}$, where $L = H_l \times W_l$ for the $l$-th decoder layer.

\begin{figure*}[t!]
    \centering
    \includegraphics[width=\textwidth]{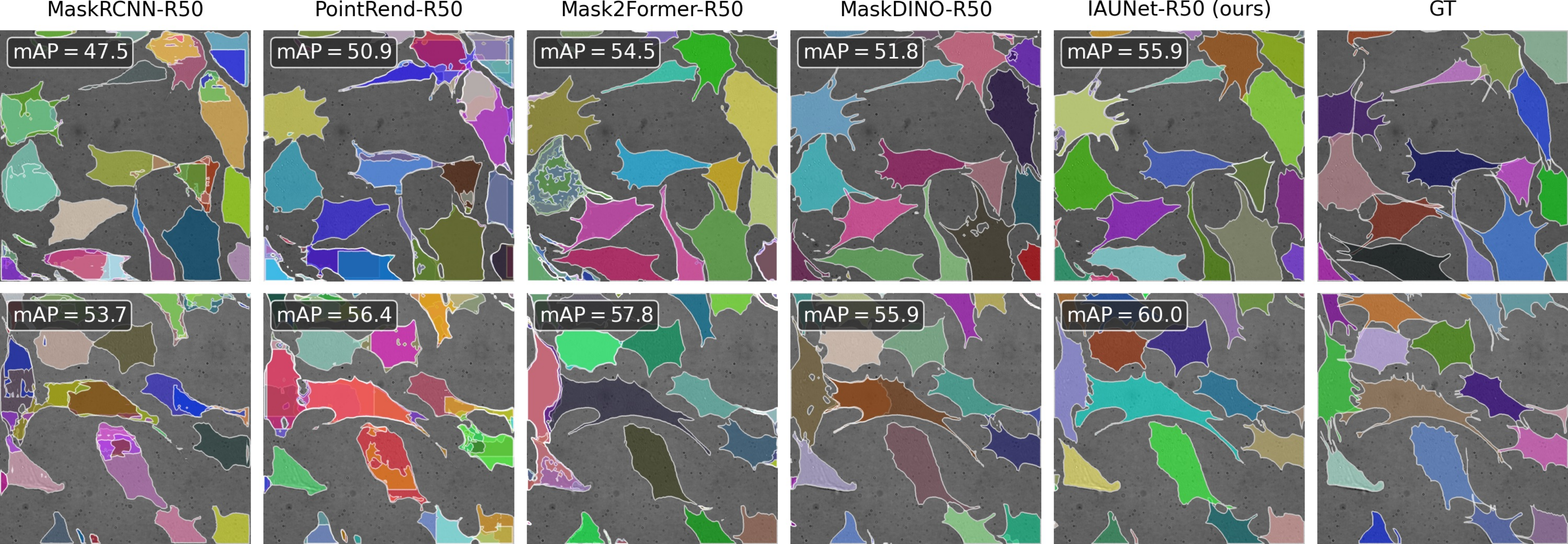}
    \caption{\textbf{Revvity-25}. Visualization of instance segmentation predictions on the Revvity-25 dataset across different state-of-the-art models (using R50 backbone). Last columns shows ground-truth annotations. IAUNet as well as MaskDINO show good generalization across tiny details and overlaping instances. We also report per-image AP score.}
    \label{fig:main_predictions_revvity_25}
\vspace{-0.3cm}
\end{figure*}

\renewcommand{\arraystretch}{1.2}
\begin{table*}[!t]
\vspace{-0.35cm}
\centering
\scriptsize
\begin{tabular}{l|c|c|p{0.4cm}p{0.4cm}|p{0.4cm}p{0.4cm}|p{0.4cm}p{0.4cm}|p{0.4cm}p{0.4cm}|p{0.4cm}p{0.4cm}|c|c}
    \multicolumn{3}{c}{} & \multicolumn{2}{c}{\textit{LIVECell}} & \multicolumn{2}{c}{\textit{EVICAN2$_E$}} & \multicolumn{2}{c}{\textit{EVICAN2$_M$}} & \multicolumn{2}{c}{\textit{EVICAN2$_D$}} & \multicolumn{2}{c}{\textit{ISBI2014}} & \multicolumn{2}{c}{} \\
    \specialrule{0.75pt}{0pt}{0pt}
    Models & backbones & num\_queries & AP & AP$_{50}$ & AP & AP$_{50}$ & AP & AP$_{50}$ & AP & AP$_{50}$ & AP & AP$_{50}$ & \#params. & FLOPs \\
    \hline
\multicolumn{15}{l}{\textit{\textbf{Models with Convolution-Based Backbones}}} \\
\hline
Mask R-CNN \cite{mask_rcnn} & R50 & 100 & \underline{44.7} & \underline{74.2} & 48.1 & 75.9 & 20.7 & 42.5 & 19.1 & 39.8 & \underline{58.9} & \textbf{88.7} & 44M & 115G \\
PointRend \cite{pointrend} & R50 & 100 & 44.0 & 73.5 & 26.6 & 47.9 & 18.0 & 38.5 & 13.4 & 28.3 & \textbf{60.0} & \textbf{88.7} & 56M & 66G \\
Mask2Former \cite{mask2former} & R50 & 100 & 43.7 & 73.8 & \underline{53.4} & \underline{89.1} & 29.1 & 54.9 & \underline{24.2} & \textbf{50.4} & 58.5 & \underline{87.5} & 44M & 67G \\
MaskDINO \cite{mask_dino} & R50 & 100 & 43.3 & 73.5 & 50.7 & 83.9 & \underline{29.3} & \underline{57.9} & 22.0 & 41.9 & 55.4 & 86.8 & 44M & 64G \\
\rowcolor{our_results_color} 
\textbf{IAUNet (ours)} & R50 & 100 & \textbf{45.3} & \textbf{75.3} & \textbf{58.0} & \textbf{91.8} & \textbf{32.1} & \textbf{59.0} & \textbf{24.9} & \underline{45.4} & 56.0 & 85.0 & 39M & 49G \\
\hline 

Mask R-CNN \cite{mask_rcnn} & R101 & 100 & \underline{44.2} & 73.2 & 41.5 & 69.9 & 23.3 & 46.9 & 17.8 & 36.7 & \textbf{60.7} & \underline{88.8} & 63M & 134G \\
PointRend \cite{pointrend} & R101 & 100 & 44.0 & \underline{73.7} & 41.3 & 65.2 & 20.2 & 39.3 & 14.8 & 32.1 & \underline{60.3} & \textbf{89.2} & 75M & 86G \\
Mask2Former \cite{mask2former} & R101 & 100 & 44.0 & 73.5 & \underline{54.4} & \underline{87.8} & 27.1 & 51.7 & 20.4 & 42.4 & 59.5 & 88.6 & 63M & 86G \\
MaskDINO \cite{mask_dino} & R101 & 100 & 43.4 & 73.6 & 53.7 & 85.0 & \underline{31.8} & \underline{59.2} & \textbf{27.1} & \textbf{51.3} & 55.7 & 87.4 & 63M & 84G \\
\rowcolor{our_results_color} 
\textbf{IAUNet (ours)} & R101 & 100 & \textbf{45.4} & \textbf{75.5} & \textbf{58.3} & \textbf{92.7} & \textbf{32.9} & \textbf{59.6} & \underline{26.9} & \underline{50.0} & 56.5 & 87.1 & 58M & 69G \\
\hline 

\multicolumn{15}{l}{\textit{\textbf{Models with Transformer-Based Backbones}}} \\
\hline
Mask R-CNN \cite{mask_rcnn} & Swin-S & 100 & 44.3 & 73.3 & 52.6 & 91.7 & 27.0 & 59.2 & 20.2 & 50.2 & \underline{61.9} & \underline{90.7} & 69M & 141G \\
PointRend \cite{pointrend} & Swin-S & 100 & 43.9 & 73.5 & 55.1 & 89.2 & 30.1 & 61.6 & 24.4 & 54.6 & \textbf{62.1} & \textbf{91.0} & 81M & 93G \\
Mask2Former \cite{mask2former} & Swin-S & 100 & 44.6 & 74.3 & \textbf{65.2} & \textbf{96.8} & \textbf{36.2} & \underline{66.7} & \textbf{30.9} & \underline{62.7} & 57.1 & 87.3 & 69M & 93G \\
MaskDINO \cite{mask_dino} & Swin-S & 100 & 43.9 & 73.8 & 57.0 & 86.9 & 33.6 & 64.9 & 27.6 & 56.9 & 52.7 & 85.3 & 71M & 181G \\
MaskDINO \cite{mask_dino} & Swin-S & 300 & 44.8 & 75.1 & 56.5 & 91.8 & \underline{35.0} & \textbf{70.7} & \underline{30.2} & \textbf{64.3} & 51.2 & 83.4 & 71M & 187G \\
\rowcolor{our_results_color} 
\textbf{IAUNet (ours)} & Swin-S & 100 & \underline{45.4} & \underline{75.4} & 58.8 & 93.1 & 32.2 & 61.9 & 27.7 & 54.1 & 61.1 & 90.1 & 64M & 76G \\
\rowcolor{our_results_color} 
\textbf{IAUNet (ours)} & Swin-S & 300 & \textbf{45.6} & \textbf{76.4} & \underline{60.9} & \underline{93.6} & 33.2 & 62.0 & 29.6 & 58.0 & 61.8 & 89.8 & 64M & 87G \\
\hline 

Mask R-CNN \cite{mask_rcnn} & Swin-B & 100 & 44.2 & 73.1 & 52.0 & 89.0 & 26.7 & 60.3 & 24.8 & 55.5 & 62.4 & \textbf{91.5} & 107M & 186G \\
PointRend \cite{pointrend} & Swin-B & 100 & 44.0 & 73.7 & 58.6 & 91.0 & 34.1 & 64.6 & 25.8 & 52.0 & \underline{62.7} & \textbf{91.5} & 119M & 137G \\
Mask2Former \cite{mask2former} & Swin-B & 100 & 44.9 & 74.7 & 55.0 & 92.5 & 31.4 & 60.9 & 27.7 & 56.6 & 58.1 & 88.4 & 107M & 138G \\
MaskDINO \cite{mask_dino} & Swin-B & 100 & 44.3 & 74.1 & 57.3 & 91.1 & 37.3 & \underline{75.7} & 30.1 & \underline{65.6} & 53.5 & 86.6 & 110M & 226G \\
MaskDINO \cite{mask_dino} & Swin-B & 300 & 45.2 & \underline{75.8} & 57.9 & 91.6 & \textbf{39.1} & \textbf{78.8} & \textbf{34.0} & \textbf{72.3} & 53.3 & 84.8 & 110M & 232G \\
\rowcolor{our_results_color} 
\textbf{IAUNet (ours)} & Swin-B & 100 & \underline{45.5} & 75.6 & \underline{59.6} & \underline{93.5} & 34.2 & 65.7 & 28.9 & 56.9 & 61.5 & \underline{90.8} & 102M & 120G \\
\rowcolor{our_results_color} 
\textbf{IAUNet (ours)} & Swin-B & 300 & \textbf{45.8} & \textbf{76.7} & \textbf{61.2} & \textbf{94.8} & \underline{38.0} & 69.6 & \underline{30.7} & 59.9 & \textbf{63.0} & \textbf{91.5} & 102M & 132G \\
\hline 

\multicolumn{15}{l}{\textit{\textbf{Specialized Cell Segmentation Methods}}} \\
\hline
CellPose \cite{cellpose} &   & - & 34.5 & 60.1 & 0.9 & 2.8 & 0.1 & 0.3 & 0.0 & 0.0 & 40.5 & 69.3 & 6.6M & 163.6G \\
CellPose + SM \cite{cellpose2} &   & - & \underline{34.9} & \underline{60.4} & \underline{8.7} & \underline{16.8} & \underline{1.6} & \underline{4.4} & \underline{2.3} & \underline{6.8} & \underline{41.6} & \underline{70.4} & 6.6M & 163.6G \\
CellDETR \cite{cell_detr} & R34 & 100 & 13.9 & 32.7 & 0 & 0.1 & 0.0 & 0.0 & 0.0 & 0.0 & 0.046 & 0.135 & 57M & 3.6T \\
\rowcolor{our_results_color} 
\textbf{IAUNet (ours)} & R50 & 100 & \textbf{45.3} & \textbf{75.3} & \textbf{58.0} & \textbf{91.8} & \textbf{32.1} & \textbf{59.0} & \textbf{24.9} & \textbf{45.4} & \textbf{56.0} & \textbf{85.0} & 39M & 49G \\
\hline 

\multicolumn{15}{l}{\textit{\textbf{YOLO Family}}} \\
\hline
YOLOv8-M \cite{yolov8} &   & - & 37.5 & 72.2 & 43.8 & 82.3 & 27.5 & 57.1 & 20.0 & 46.2 & 54.9 & 90.7 & 27.2M & 110.4G \\
YOLOv8-L \cite{yolov8} &   & - & 40.5 & 72.5 & 44.7 & 83.1 & 28.1 & 58.2 & 20.3 & 46.1 & 55.1 & \underline{91.1} & 45.9M & 220.8G \\
YOLOv8-X \cite{yolov8} &   & - & \underline{41.1} & \underline{73.1} & \underline{45.8} & \underline{85.6} & \underline{28.9} & \underline{59.2} & \underline{20.7} & \underline{47.3} & \underline{55.3} & \textbf{91.4} & 71.8M & 344.5G \\
\rowcolor{our_results_color} 
\textbf{IAUNet (ours)} & Swin-S & 100 & \textbf{45.4} & \textbf{75.4} & \textbf{58.8} & \textbf{93.1} & \textbf{32.2} & \textbf{61.9} & \textbf{27.7} & \textbf{54.1} & \textbf{61.1} & 90.1 & 64M & 76G \\
\hline 

YOLOv9-E \cite{yolov9} &   & - & 41.2 & \underline{73.2} & 45.6 & 84.4 & 27.2 & 57.9 & 20.1 & 47.3 & 53.3 & \textbf{91.6} & 27.8M & 159.1G \\
YOLOv9-C \cite{yolov9} &   & - & \underline{41.4} & 73.1 & \underline{45.9} & \underline{85.6} & \underline{28.3} & \underline{59.8} & \underline{22.2} & \underline{49.9} & \underline{55.7} & \underline{91.1} & 60.5M & 248.1G \\
\rowcolor{our_results_color} 
\textbf{IAUNet (ours)} & Swin-S & 100 & \textbf{45.4} & \textbf{75.4} & \textbf{58.8} & \textbf{93.1} & \textbf{32.2} & \textbf{61.9} & \textbf{27.7} & \textbf{54.1} & \textbf{61.1} & 90.1 & 64M & 76G \\
\hline 

\multicolumn{15}{l}{\textit{\textbf{SAM Family}}} \\
\hline
SAM-B \textit{(points)} \cite{sam} &   & - & 5.0 & 12.4 & 28.4 & 56.0 & 5.4 & 13.8 & 3.2 & 7.2 & 33.8 & 51.8 & 90M & 742G \\
SAM-B \textit{(boxes)} \cite{sam} &   & - & \underline{24.3} & \underline{56.9} & \underline{55.0} & \textbf{96.6} & \textbf{38.6} & \textbf{91.2} & \textbf{34.8} & \textbf{82.3} & \underline{59.6} & \textbf{92.8} & 90M & 742G \\
\rowcolor{our_results_color} 
\textbf{IAUNet (ours)} & Swin-S & 100 & \textbf{45.4} & \textbf{75.4} & \textbf{58.8} & \underline{93.1} & \underline{32.2} & \underline{61.9} & \underline{27.7} & \underline{54.1} & \textbf{61.1} & \underline{90.1} & 64M & 76G \\
\hline 

SAM-L \textit{(points)} \cite{sam} &   & - & 6.3 & 13.6 & 28.1 & 54.1 & 4.9 & 12.4 & 3.2 & 7.5 & 32.8 & 51.0 & 308M & 2.6T \\
SAM-L \textit{(boxes)} \cite{sam} &   & - & \underline{29.2} & \underline{65.2} & \underline{57.2} & \textbf{96.6} & \textbf{45.8} & \textbf{95.3} & \textbf{39.7} & \textbf{88.6} & \underline{60.8} & \textbf{93.6} & 308M & 2.6T \\
\rowcolor{our_results_color} 
\textbf{IAUNet (ours)} & Swin-B & 300 & \textbf{45.8} & \textbf{76.7} & \textbf{61.2} & \underline{94.8} & \underline{38.0} & \underline{69.6} & \underline{30.7} & \underline{59.9} & \textbf{63.0} & \underline{91.5} & 102M & 132G \\
\hline

 \end{tabular}
 \caption{\textbf{Instance segmentation on LIVECell, EVICAN2 (Easy, Medium, Difficult), and ISBI2014.} IAUNet outperforms strong query-based Mask2Former and MaskDINO baselines for both AP and AP$_{50}$ when training with fewer parameters. For a fair comparison, we only consider single-scale inference and models trained until full convergence. IAUNet remains efficient across different backbones.}
 \label{table:metrics_all}
 \vspace{-0.4cm}
\end{table*}

\subsubsection{Positional Embeddings}
\label{sec:positional_embeddings}
To maintain spatial awareness, which is crucial for Transformer-based models, we add $N$ learnable positional embeddings to both instance queries. Following the previous work \cite{detr}, we add sinusoidal positional embeddings $e_{pos} \in \mathbb{R}^{H_lW_l \times D}$ to the mask $X_m$.

\subsubsection{Instance Queries Update}
\label{sec:instance_queries_update}

We update $N$ instance queries with the mask features $X_{m}$ using the cross-attention layer (\cref{fig:iaunet_main}, red block) followed by the self-attention layer between queries and FFN layer. Thus, all queries attend to each other, ensuring better object separation. The update is expressed as follows: 

\vspace{-0.4cm}
\begin{align}
\hat{X}_{l} &= \text{softmax}\left(Q_{l} K_{l}^T\right) V_{l} + X_{l-1} \\ 
X_{l} &= \text{FFN}(\hat{X}_{l})
\end{align}

where \( Q_{l} = f_Q(q_{l}) \in \mathbb{R}^{N \times 256} \) represents the transformed queries at layer \( l \), and the keys and values \( K_{l}, V_{l} \in \mathbb{R}^{H_l W_l \times 256} \) are computed from the mask features $X_m$. The queries are updated sequentially within Transformer blocks at each decoder layer.

\subsubsection{Mask Head}
\label{sec:mask_head}

To keep the prediction process lightweight without performance loss, we fuse only high-resolution features. As shown in (\cref{fig:iaunet_main}, red arrows), we construct a pixel embedding map by combining the 1/4 resolution backbone feature map $X_b$ with an upsampled 1/8 resolution mask features $X_m$ from the Pixel decoder. Specifically, we apply two linear projections on the refined instance queries $q$ to obtain mask embeddings $q_c$ and object class scores. The final mask prediction is obtained by taking the dot product of each mask embedding with this fused feature map: 

\vspace{-0.4cm}
\begin{align}  
m = q_c \otimes \mathcal{M} \left( \mathcal{F}(X_b) + \mathcal{U}(X_m) \right),  
\end{align}  

where $\mathcal{M}$ is the segmentation head, $\mathcal{F}$ is a convolutional layer that adjusts the channel dimensions to match the Transformer hidden space, and $\mathcal{U}$ is a simple $2\times$ upsampling function applied to $X_m$. Besides, each instance query predicts the object class probability, including a "no object" ($\varnothing$). During inference, we re-score the predicted masks. For each instance, we calculate the maskness metric \cite{mask2former}, denoted as $p_i = \frac{1}{N} \sum_{i=1}^{N} m_i$, where $m \in \{M_n\}_{n=1}^N$ is the predicted instance mask. The combined confidence score for each instance is then computed by multiplying the class probability score $c_i$ with the maskness score $p_i$: $\hat{c}_i = c_i \cdot p_i$.

\subsection{Mask Level Matching}
\label{mask_level_matching}

During training, the model outputs \( \{M_n\}_{n=1}^N \) predicted masks, where \( N > M \), the number of ground truth masks \( \{G_k\}_{k=1}^M \). To compute losses on matched predictions, we perform bipartite matching between \( \{M_n\} \) and \( \{G_k\} \) using the Hungarian algorithm \cite{hungarian_algorithm}, which finds the optimal permutation \( \sigma \) that minimizes the matching cost:
\begin{equation}
\sigma = \arg\min_{\sigma \in S} \sum_{i=1}^{M} \mathcal{L}_{\text{match}}(M_{\sigma(i)}, G_i).
\end{equation}
For the matching cost, we use a combination of classification and mask costs:
\begin{equation}
\mathcal{L}_{\text{match}} = \lambda_{\text{cls}} \cdot \mathcal{L}_{\text{cls}} + \lambda_{\text{dice}} \cdot \mathcal{L}_{\text{dice}} + \lambda_{\text{bce}} \cdot \mathcal{L}_{\text{bce}}
\end{equation}

Following \cite{mask2former} we set \( \lambda_{\text{cls}} = 1.0 \), \( \lambda_{\text{dice}} = 2.0 \), and \( \lambda_{\text{bce}} = 5.0 \) to control the weight of each cost term. Here, \( \mathcal{L}_{\text{cls}} \) represents the cross-entropy loss for object classification, with a "no object" class weighted at 0.1. The terms \( \mathcal{L}_{\text{bce}} \) and \( \mathcal{L}_{\text{dice}} \) denote the binary cross-entropy loss and Dice loss, respectively, for the segmentation masks \cite{vnet}.

For the loss function, we align it with the matching cost by applying the same coefficients to ensure consistency. The final loss function is defined as:
\begin{equation}
\mathcal{L} = \lambda_{\text{cls}} \cdot \mathcal{L}_{\text{cls}} + \lambda_{\text{dice}} \cdot \mathcal{L}_{\text{dice}} + \lambda_{\text{bce}} \cdot \mathcal{L}_{\text{bce}}
\end{equation}

\section{Experiments}
\label{sec:experiments}

In this section, we evaluate our IAUNet on multiple datasets, including our novel Revvity-25 dataset. We also compare it with multiple state-of-the-art models in terms of segmentation performance. Besides, we conduct ablation studies and show the effectiveness of our model components. To provide a comprehensive comparison, we use a range of datasets:

\noindent \textbf{LIVECell \cite{livecell}} 
\label{par:livecell}
is one of the most extensive datasets regarding images and annotated cells for instance segmentation. It consists of 5,239 high-resolution phase-contrast images (520×704 pixels) with over 1.6 million expert-validated annotated cells. It includes eight cell types with varied shapes and densities.

\noindent \textbf{EVICAN2 \cite{evican}} 
\label{par:evican}
is the most heterogeneous dataset for cell segmentation, containing 5,237 microscopy images across brightfield, phase contrast, and fluorescence modalities, with 52,959 annotated cell and nucleus instances. It includes training and validation sets with 4,640 partially annotated images and a test set of 98 fully annotated images. The test set is categorized by difficulty based on image quality: easy, medium, and difficult.

\renewcommand{\arraystretch}{1.2}
\begin{table*}[!t]
\vspace{-0.35cm}
\centering
\scriptsize
\begin{tabular}{l|c|c|p{0.4cm}p{0.4cm}p{0.4cm}|p{0.4cm}p{0.4cm}p{0.4cm}|c|c}
    \multicolumn{3}{c}{} & \multicolumn{2}{c}{\textit{Revvity-25}} & \multicolumn{6}{c}{} \\
    \specialrule{0.75pt}{0pt}{0pt}
    Models & backbones & num\_queries & AP & AP$_{50}$ & AP$_{75}$ & AP$_{S}$ & AP$_{M}$ & AP$_{L}$ & \#params. & FLOPs \\
    \hline
\multicolumn{7}{l}{\textit{\textbf{Models with Convolution-Based Backbones}}} \\
\hline
Mask R-CNN \cite{mask_rcnn} & R50 & 100 & 39.7 & 77.2 & 37.4 & 0.6 & 19.0 & 44.6 & 44M & 115G \\
PointRend \cite{pointrend} & R50 & 100 & 42.2 & 79.4 & 40.9 & 0.4 & 21.7 & 47.3 & 56M & 66G \\
Mask2Former \cite{mask2former} & R50 & 100 & \underline{46.4} & 79.8 & \underline{49.9} & \underline{0.7} & \underline{25.7} & \underline{52.8} & 44M & 67G \\
MaskDINO \cite{mask_dino} & R50 & 100 & 45.6 & \underline{80.4} & 48.2 & \textbf{1.8} & 22.3 & 51.8 & 44M & 64G \\
\rowcolor{our_results_color} 
\textbf{IAUNet (ours)} & R50 & 100 & \textbf{49.7} & \textbf{82.1} & \textbf{54.8} & 0.6 & \textbf{27.3} & \textbf{56.0} & 39M & 49G \\
\hline 

Mask R-CNN \cite{mask_rcnn} & R101 & 100 & 40.7 & 77.5 & 39.9 & 0.4 & 20.1 & 45.8 & 63M & 134G \\
PointRend \cite{pointrend} & R101 & 100 & 42.9 & 79.3 & 42.5 & 0.0 & 18.4 & 48.9 & 75M & 86G \\
Mask2Former \cite{mask2former} & R101 & 100 & 47.2 & 80.1 & \underline{51.8} & \textbf{1.7} & \underline{25.7} & 53.3 & 63M & 86G \\
MaskDINO \cite{mask_dino} & R101 & 100 & \underline{47.3} & \underline{81.0} & 50.4 & \underline{0.9} & 23.0 & \underline{53.5} & 63M & 84G \\
\rowcolor{our_results_color} 
\textbf{IAUNet (ours)} & R101 & 100 & \textbf{51.5} & \textbf{84.7} & \textbf{56.1} & \textbf{1.7} & \textbf{29.2} & \textbf{57.8} & 58M & 69G \\
\hline 

\multicolumn{7}{l}{\textit{\textbf{Models with Transformer-Based Backbones}}} \\
\hline
Mask R-CNN \cite{mask_rcnn} & Swin-S & 100 & 24.7 & 63.4 & 12.5 & 0.0 & 7.3 & 28.9 & 69M & 141G \\
PointRend \cite{pointrend} & Swin-S & 100 & 43.6 & 80.0 & 43.0 & 0.5 & 21.5 & 48.9 & 81M & 93G \\
Mask2Former \cite{mask2former} & Swin-S & 100 & 51.2 & 83.3 & 56.4 & 2.7 & 27.7 & 58.0 & 69M & 93G \\
MaskDINO \cite{mask_dino} & Swin-S & 100 & 50.3 & 83.2 & 53.9 & \textbf{4.7} & 27.6 & 56.1 & 71M & 181G \\
MaskDINO \cite{mask_dino} & Swin-S & 300 & 49.4 & 83.6 & 53.3 & \underline{2.9} & 25.8 & 55.3 & 71M & 187G \\
\rowcolor{our_results_color} 
\textbf{IAUNet (ours)} & Swin-S & 100 & \underline{53.0} & \underline{85.7} & \underline{57.0} & 1.3 & \textbf{29.7} & \underline{59.1} & 64M & 76G \\
\rowcolor{our_results_color} 
\textbf{IAUNet (ours)} & Swin-S & 300 & \textbf{53.3} & \textbf{86.0} & \textbf{59.6} & 1.6 & \underline{29.4} & \textbf{59.8} & 64M & 87G \\
\hline 

Mask R-CNN \cite{mask_rcnn} & Swin-B & 100 & 27.1 & 64.9 & 17.2 & 0.1 & 9.7 & 31.2 & 107M & 186G \\
PointRend \cite{pointrend} & Swin-B & 100 & 45.2 & 80.1 & 47.9 & 0.1 & 23.0 & 50.9 & 119M & 137G \\
Mask2Former \cite{mask2former} & Swin-B & 100 & 52.0 & 83.6 & \underline{58.4} & \underline{1.1} & 27.8 & 59.0 & 107M & 138G \\
MaskDINO \cite{mask_dino} & Swin-B & 100 & 50.5 & 83.5 & 54.9 & \textbf{2.0} & 27.1 & 56.4 & 110M & 226G \\
MaskDINO \cite{mask_dino} & Swin-B & 300 & 50.4 & 84.3 & 54.8 & 0.8 & 26.3 & 56.6 & 110M & 232G \\
\rowcolor{our_results_color} 
\textbf{IAUNet (ours)} & Swin-B & 100 & \underline{53.5} & \underline{86.1} & \textbf{59.4} & 0.8 & \textbf{30.5} & \underline{59.7} & 102M & 120G \\
\rowcolor{our_results_color} 
\textbf{IAUNet (ours)} & Swin-B & 300 & \textbf{53.7} & \textbf{86.5} & \textbf{59.4} & 1.0 & \underline{30.0} & \textbf{60.3} & 102M & 132G \\
\hline

 \end{tabular}
 \caption{\textbf{Instance segmentation on our Revvity-25 dataset.} IAUNet outperforms strong query-based Mask2Former and MaskDINO baselines as well as other state-of-the-art models when training with fewer parameters. For a fair comparison, we only consider single-scale inference and models trained until full convergence. IAUNet also efficiently scales with more queries while remaining efficient.}
 \label{table:metrics_all_revvity}
 \vspace{-0.3cm}
\end{table*}

\noindent \textbf{ISBI2014 \cite{isbi2014}} 
\label{par:isbi2014}
is a dataset from the Overlapping Cervical Cytology Image Segmentation Challenge. It includes 16 real extended depth-of-focus (EDF) cervical cytology images and 945 synthetic images. The dataset provides high-quality pixel-level annotations for nuclei and cytoplasm, with a resolution of 512 $\times$ 512. We follow the challenge setting \cite{isbi2014}, using 45 synthetic images for training, 90 for validation, and 810 for testing.

One of our key contributions in this paper is a novel cell instance segmentation dataset named \textbf{Revvity-25}. It includes 110 high-resolution 1080 $\times$ 1080 brightfield images, each containing, on average, 27 manually labeled and expert-validated cancer cells, totaling 2937 annotated cells. To our knowledge, this is the first dataset with accurate and detailed annotations for cell borders and overlaps, with each cell annotated using an average of 60 polygon points, reaching up to 400 points for more complex structures. Revvity-25 dataset provides a unique resource that opens new possibilities for testing and benchmarking models for modal and amodal semantic and instance segmentation.

\subsection{Implementation Details}
\label{sec:implementation_details}
All experiments were conducted on a single Tesla V100 GPU with 32GB memory. We adopt the training scheme published in earlier works \cite{mask2former}. We use the CosineAnnealingLR scheduler \cite{cosine_annealing_lr} with a minimum learning rate of 1e-6, and the AdamW optimizer \cite{adamw} with an initial learning rate of 1e-4 and weight decay of 0.05. During training, we employ longest-side resizing to scale all images to 512 $\times$ 512 pixels, preserving the original aspect ratio. For augmentation, we apply scale jittering \cite{copy-paste} within a scale range of 0.8 to 1.5, followed by fixed-size cropping to 512 $\times$ 512 and random flipping. All models were trained to full convergence with a batch size of 8. Unless specified, we apply the same resizing process during inference, using a consistent mask prediction threshold of 0.5 across all models.

\subsection{Main Results}
\label{sec:main_results}

In this section, we outline the dataset setup for training and present the results. For the LIVECell dataset, we preprocess images by randomly cropping them to a maximum of 100 instances, ensuring consistency in prediction counts across datasets. We use the original train, validation, and test splits for all models. For the ISBI2014 dataset, we follow the original train, validation, and test splits. All models, except CellPose \cite{cellpose}, are trained to segment both cell and nuclei classes. Since CellPose does not support multi-class segmentation by default, we train separate models for each class and average the performance. The Revvity-25 dataset is divided equally into train and test sets, each containing 55 images. For EVICAN2, we report results on the easy, medium, and difficult test sets. A maximum of 100 queries is set across all datasets. For example, in \cref{table:metrics_all}, IAUNet is compared with state-of-the-art models across diverse datasets. In models with convolution-based backbones, IAUNet with ResNet-50 achieves an AP of 45.3 and AP\(_{50}\) of 75.3 on LiveCell. It outperforms Mask R-CNN, PointRend, Mask2Former, and MaskDINO while using fewer parameters (39M) and lower FLOPs (49G). With a ResNet-101 backbone, IAUNet records an AP of 45.4 and AP\(_{50}\) of 75.5. IAUNet also scales better compared to MaskDINO when using transformer-based backbones. While IAUNet performs best on LIVECell but has room for improvement on ISBI2014, where the low object count leads to some queries predicting duplicates. Among specialized cell segmentation methods, IAUNet outperforms CellPose, CellPose + SM, and CellDETR. CellDETR, scaled to 100 objects with a softmax head on high-resolution images, has high computational cost and parameter count, making it unsuitable for some datasets. CellPose struggles to generalize when object sizes differ significantly between train and test sets, as seen in EVICAN2, due to its reliance on object diameter for post-processing.

In \cref{fig:main_predictions_revvity_25}, we visualize the predictions and compute an image-wise AP score. IAUNet consistently outperforms other state-of-the-art models. IAUNet visibly offers more detailed segmentation, capturing longer pixel relationships and effectively handling overlapping regions in some cases. In \cref{table:metrics_all_revvity}, we demonstrate IAUNet’s strengths on the Revvity-25 dataset, where it achieves the highest scores across multiple backbones, with an AP of 49.7 using ResNet-50 and 53.7 with Swin-B.

\renewcommand{\arraystretch}{1.2}
\begin{table}[!t]
\vspace{-0.35cm}
\centering
\scriptsize
\begin{tabular}{l|p{0.4cm}p{0.4cm}p{0.4cm}|c}
    \multicolumn{1}{c}{} & \multicolumn{3}{c}{\textit{}} & \multicolumn{1}{c}{} \\
    Pixel Decoder & AP & AP$_{50}$ & AP$_{75}$ & FLOPs \\
    \hline
\hspace{0.5em}+ full skip & \textbf{44.7} & \textbf{73.9} & \textbf{48.9} & 146G \\
\hspace{0.5em}+ $1 \times 1$ skip concat & 44.2 & \underline{73.8} & \underline{48.3} & 135G \\
\hspace{0.5em}+ $1 \times 1$ skip add & \underline{44.3} & 73.3 & 48.2 & 132G \\
\hspace{0.5em}+ light mask head & 43.8 & 73.1 & 47.4 & 42G \\
 \end{tabular}
 \caption{\textbf{Pixel Decoder Variants (Skip Connections).} We retain skip connection concatenation as in \cref{eq:main_feats_update} and introduce a lightweight mask head.}
 \label{table:ablation_pixel_decoder_skips}
 \vspace{-0.3cm}
\end{table}

\subsection{Ablation Studies}
\label{sec:ablation_studies}
In this section, we present an ablation study to evaluate the impact of each component in our model architecture. We focus on analyzing the contributions of the Pixel decoder and the Transformer decoder to overall model performance. All ablation studies were conducted on the LIVECell dataset. 

\noindent \textbf{Skip Connections.} IAUNet builds on the U-Net architecture. \cref{table:ablation_pixel_decoder_skips} presents the impact of different skip connection configurations. The model performs best with full skip connections over main features $X$, where channels are not reduced. To balance computational efficiency, skip channels are reduced to 256 via $1 \times 1$ convolutions before fusing features using concatenation or addition. Concatenation produces optimal performance and stability, while addition creates an FPN-like \cite{fpn} structure in the decoder with a further performance drop. Finally, adding a light mask head to produce high-resolution features further reduces the FLOP count to 42G without a significant performance drop.

\noindent \textbf{Pixel Decoder.} In \cref{table:ablation_pixel_decoder_components}, we study each component of the Pixel decoder separately. To further refine features in the Pixel decoder, decoupling mask features with a dedicated mask branch helps. To improve scalability, we reduce the feed-forward dimension to 1024 and add a Squeeze-and-Excitation \cite{squeeze_and_excitation} block to enhance feature representation. We observe that the model benefits from additional spatial information for multiple grouped objects of irregular shapes. Using CoordConv \cite{coord_conv} at each level enriches the main features $X$ before further processing, helping the model better capture object locations and improve translation awareness. This modification improves segmentation performance, increasing AP to 44.7.

\noindent \textbf{Transformer Decoder.} 
We evaluate the impact of scaling the Transformer decoder in \cref{table:ablation_pixel_decoder_components}. First, we introduce three Transformer decoder blocks per decoder layer, resulting in a total of \( 3L \) Transformer blocks. We explore two main strategies for refining object queries. The first approach, inspired by \cite{mask2former}, follows a Round-Robin cycle update, where queries are refined in one Transformer block from each decoder layer at a time and passed to the next, forming a cycle that returns to low-resolution features. In contrast, we propose a sequential (seq.) update strategy, where object queries are refined within all decoder blocks per decoder layer first, increasing AP to 45.1. 
Building on this, we apply deep supervision by computing the loss after each Transformer decoder layer using the updated queries and high-resolution Pixel decoder features $X_m$.
Additionally, in \cref{table:ablation_num_queries}, we evaluate the scalability of the number of queries, showing that the model achieves peak performance as the number of queries increases.

\renewcommand{\arraystretch}{1.2}
\begin{table}[!t]
\vspace{-0.35cm}
\centering
\scriptsize
\begin{tabular}{l|p{0.4cm}p{0.4cm}p{0.4cm}|c|c}
    \multicolumn{1}{c}{} & \multicolumn{3}{c}{\textit{}} & \multicolumn{2}{c}{} \\
    Decoder & AP & AP$_{50}$ & AP$_{75}$ & \#params. & FLOPs \\
    \hline
\textbf{IAUNet (R50)} & 43.8 & 73.1 & 47.4 & 34M & 42G \\
\hline
\hspace{0.5em}+ mask branch $X_m$ & 44.0 & 73.2 & 47.9 & 34M & 42G \\
\hspace{0.5em}+ FFN (2048 $\rightarrow$ 1024) & 44.1 & 73.2 & 48.0 & 32M & 42G \\
\hspace{0.5em}+ SE block \cite{squeeze_and_excitation} & 44.2 & 73.3 & 48.1 & 32M & 42G \\
\hspace{0.5em}+ CoordConv \cite{coord_conv} & 44.7 & 74.1 & \underline{48.7} & 32M & 42G \\

\hline

\hspace{0.5em}+ $L$ (1 $\rightarrow$ 3) (cycle.) & 44.3 & 74.0 & 48.1 & 39M & 49G \\
\hspace{0.5em}+ $L$ (1 $\rightarrow$ 3) (seq.) & \underline{45.1} & \underline{74.4} & \textbf{49.4} & 39M & 49G \\
\hspace{0.5em}+ deep\_supervision & \textbf{45.3} & \textbf{75.3} & \textbf{49.4} & 39M & 49G \\

 \end{tabular}
 \caption{\textbf{Decoder}. We investigate the benefit of adding different decoder components. Adding CoordConv \cite{coord_conv} improves object localization. Scaling the Transformer decoder with deep supervision shows best performance.}
 \label{table:ablation_pixel_decoder_components}
 \vspace{-0.3cm}
\end{table}

\renewcommand{\arraystretch}{1.2}
\begin{table}[!t]
\vspace{-0.35cm}
\centering
\scriptsize
\begin{tabular}{c|p{0.4cm}p{0.4cm}p{0.4cm}|c}
    \multicolumn{3}{c}{\textit{}} & \multicolumn{2}{c}{} \\
    num\_queries & AP & AP$_{50}$ & AP$_{75}$ & FLOPs \\
    \hline
100 & 45.3 & 75.3 & 49.4 & 49G \\
300 & \underline{45.9} & \underline{76.5} & \underline{50.4} & 61G \\
500 & \textbf{46.1} & \textbf{76.8} & \textbf{50.8} & 73G \\
1000 & 45.3 & 76.3 & 50.0 & 104G \\

 \end{tabular}
 \caption{\textbf{Num. queries}. Scaling the number of object queries benefits the model.}
 \label{table:ablation_num_queries}
 \vspace{-0.3cm}
\end{table}

\vspace{-0.15cm}

\section{Conclusions}
\label{sec:conclusions}

We introduce IAUNet, a novel query-based U-Net architecture with a lightweight convolutional Pixel decoder and a Transformer decoder that supervises object-specific queries for instance segmentation in biomedical imaging. Our model outperforms leading methods, particularly for medium and large objects, and sets a strong baseline for cell segmentation tasks, as demonstrated on our Revvity-25 Dataset. 
While IAUNet performs well in most tasks, it struggles with small object segmentation and could benefit from optimization for high-instance images. Future work will focus on improving performance in these areas.

\section{Acknowledgments}
\label{sec:acknowledgments}

This work was supported by Revvity and funded by the TEM-TA101 grant “Artificial Intelligence for Smart Automation.” Computational resources were provided by the High-Performance Computing Cluster at the University of Tartu. We thank the Biomedical Computer Vision Lab for their invaluable support. We express gratitude to the Armed Forces of Ukraine and the bravery of the Ukrainian people for enabling a secure working environment, without which this work would not have been possible.

{
    \small
    \bibliographystyle{ieeenat_fullname}
    \bibliography{main}
}

\end{document}